\documentclass[letterpaper, 10 pt, conference]{ieeeconf}  %
\IEEEoverridecommandlockouts  %
\usepackage{times}
\usepackage{multicol}
\usepackage[bookmarks=true,colorlinks]{hyperref}
\usepackage{graphicx}
\usepackage{etoolbox}
\usepackage[noadjust]{cite}
\usepackage{amsmath}
\usepackage{amssymb}  %
\usepackage{booktabs}
\usepackage{multirow}
\usepackage{algorithm}
\usepackage{algpseudocode}
\usepackage{svg}
\usepackage{adjustbox}
\usepackage{subfig}
\usepackage{color,xcolor}

\newcommand{\insertfig}{
    \includegraphics[width=\textwidth,trim={5.7cm, 2.5cm, 6.8cm, 2cm},clip]{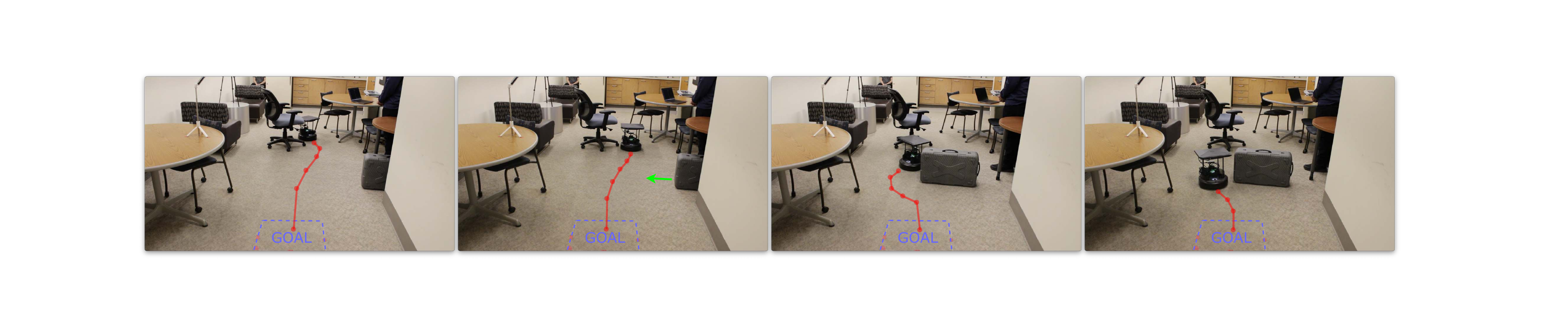}\captionof{figure}{\label{fig:teaser}
        We present a framework for trajectory optimization to navigate mobile robots in dynamic indoor environments.
        Our proposal uses robot-centric RGB-D information combined with prior knowledge to infer composite SDF-based representations, which are queried to obtain collision costs and gradients to generate collision-free trajectories.
        As shown above, the robot initially plans a trajectory to the goal region while avoiding the office chair (\emph{left} and \emph{left-middle}).
        Subsequently, a briefcase is placed in front of the robot as an obstruction (\emph{right-middle}), which the robot dynamically avoids to proceed towards the goal position.
    }\vspace{-2mm}
}

\makeatletter
\apptocmd{\@maketitle}{\centering\insertfig}{}{}%
\makeatother

\begin{document}

\title{\LARGE\bf
    Differentiable Composite Neural Signed Distance Fields for Robot Navigation in Dynamic Indoor Environments
}

\author{S. Talha Bukhari, Daniel Lawson, and Ahmed H. Qureshi%
    \thanks{
        S. Talha Bukhari, Daniel Lawson, and Ahmed H. Qureshi are with the Department of Computer Science, Purdue University, West Lafayette, IN 47907, USA
        {\tt\small \{bukhars,lawson95,ahqureshi\}@purdue.edu}
    }%
}

\maketitle

\begin{abstract}
Neural Signed Distance Fields (SDFs) provide a differentiable environment representation to readily obtain collision checks and well-defined gradients for robot navigation tasks.
However, updating neural SDFs as the scene evolves entails re-training, which is tedious, time consuming, and inefficient, making it unsuitable for robot navigation with limited field-of-view in dynamic environments.
Towards this objective, we propose a compositional framework of neural SDFs to solve robot navigation in indoor environments using only an onboard RGB-D sensor.
Our framework embodies a dual mode procedure for trajectory optimization, with different modes using complementary methods of modeling collision costs and collision avoidance gradients.
The primary stage queries the robot body's SDF, swept along the route to goal, at the obstacle point cloud, enabling swift local optimization of trajectories.
The secondary stage infers the visible scene's SDF by aligning and composing the SDF representations of its constituents, providing better informed costs and gradients for trajectory optimization.
The dual mode procedure combines the best of both stages, achieving a success rate of 98\%, 14.4\% higher than baseline with comparable amortized plan time on iGibson 2.0.
We also demonstrate its effectiveness in adapting to real-world indoor scenarios.
The video demonstrations and code are available at the \url{https://stalhabukhari.github.io/icra25-sdf-dyn-nav}.
\end{abstract}

\section{Introduction}

Indoor robot navigation with local onboard sensors in dynamic environments is a challenging problem.
Most existing solutions relying on unrealistic assumptions such as access to a known or trackable trajectory of moving obstacles~\cite{finean2021predicted}.
In practice, these assumptions can only be fulfilled with external overhead cameras, which is not pragmatic given the overarching goal of building intelligent service robots which can operate in a self-contained manner with onboard sensors in real life scenarios.
Such robot systems will enable a broad set of applications ranging from home assistance to warehouse management.
For instance, assistive robots in home environments will have to navigate around chairs or toys, which are often relocated arbitrarily. 
Similarly, in warehouses the objects are moved around for scheduling, and the robots retrieving them must safely navigate arbitrary scenarios.
Despite large-scale applications, such an essential building block of intelligent systems remains an unfulfilled challenge.

To navigate dynamic environments from local observations, it is critical to form a rich representation of obstacles for collision avoidance.
In this regard, Signed Distance Fields (SDFs) provide a useful representation framework with well-defined gradients that are useful for solving robot motion optimization under collision avoidance constraints~\cite{zucker2013chomp,oleynikova2017voxblox,finean2021predicted}.
Recent advancements have led to deep learning-based SDF modeling paradigms~\cite{deepsdf19,isdf22,gropp2020implicit}, which allow real-time inference at arbitrary resolutions, along with gradients available upfront via automatic differentiation.
From the robot navigation perspective, neural SDFs have been employed for trajectory optimization~\cite{saulnier2020information,hiosdf2023,nfomp2022}.
However, these approaches either assume static obstacles or full observability of the environment, making them unsuitable for the proposed robot navigation task.
Hence, modeling SDF of a dynamic environment using neural approaches is relatively unexplored.

Towards this objective, we present a novel approach for solving robot navigation in dynamic indoor environments by inferring neural SDFs using only local onboard sensors.
Instead of modeling the environment from the ground up as the robot navigates, we combine local sensory information with prior knowledge to compute the collision costs and gradients serving trajectory optimization.
Our proposal encompasses the following contributions:
\begin{enumerate}
    \item A compositional framework of neural SDFs, inferring the visible scene's SDF using only robot-centeric visual input, to generate trajectories optimized for collisions and effort.
    \item A simple, yet effective trajectory optimization procedure using the robot body's SDF representation to swiftly query collision costs and collision avoidance gradients.
    \item A novel two-stage algorithm which enables a better trade-off between quick inference and accurate gradient information to guide trajectory optimization.
\end{enumerate}

\section{Related Work}

Robot navigation in indoor environments has been studied over the past three decades~\cite{desouza2002vision, cox1991blanche}, with most approaches falling under classical, imitation-learning (IL), and reinforcement-learning (RL) paradigms.
Classical methods mostly assume a known map, and solve navigation using traditional robot motion planning methods \cite{burgard1999experiences, thrun1999minerva, karaman2011sampling}.
Other classical approaches that scale to unknown maps rely on Simultaneous Localization and Mapping (SLAM) in the loop \cite{jones2011visual, chaplot2020learning, zhan2022activermap}.
However, classical approaches usually suffer from large computational times, making them less lucrative for real-time applications.
Advancements in machine learning approaches resulted in the emergence of IL and RL-based methods. 
IL methods require an expert in the loop or its demonstration data for learning the navigation policies \cite{gupta2017cognitive, finn2017one, stepputtis2020language}.
In contrast, RL-based techniques learn by trial-and-error through interaction with the environment \cite{zhu2017target, mirowski2016learning, ye2021auxiliary}.
However, both IL and RL methods face challenges in translation to practical scenarios due to the need for massive amounts of diverse data and a lack of interpretability.
Instead, we present a principled, data-efficient approach based on implicit neural representations to model the environment for collision avoidance.

The representation framework used to model obstacles in the environment is an important factor in devising efficient motion planning algorithms.
In this regard, a number of different environment representation methods have been used~\cite{oleynikova2016signed, mildenhall2021nerf, nfomp2022, ni2022ntfields}, the signed distance field (SDF) being the most prominent and widely used in robotics research.
Recent advancements have led to faster ways of computing SDF via truncation and sparse voxelization \cite{pan2022voxfield, oleynikova2017voxblox, han2019fiesta}, leading to various approaches that leverage them for robot navigation in static environments.
Since computing the SDF of the full environment is not necessary for trajectory optimization, EGO-Planner \cite{zhou2020ego} formulates an approximation of the SDF, limiting the amount of environment information being considered.
However, its experimental setup involves simple geometries such as thin cylinders, which do not provide significant occlusion.
On the other hand, iPlanner~\cite{yang2023iplanner} directly infers trajectories from depth images by implicitly learning the SDF-guided trajectory optimization procedure.
A different line of work considers modeling the environment through Neural Radiance Fields (NeRFs) wherein the density estimates provide a proxy for workspace occupancy~\cite{adamkiewicz2022vision}.
However, NeRFs are optimized for novel view synthesis and do not accurately capture object surface information used to delineate formidable and free workspace regions and derive smooth gradients.

Recent approaches also consider navigating dynamic environments using SDF-based obstacle representations.
For instance, \cite{finean2021predicted} uses SDFs for trajectory optimization to find a path for a robot manipulator to maneuver around a moving obstacle.
However, the approach considers simple obstacle geometry with trackable linear motion, whereas daily-life objects have intricate shapes and geometries and are often not trackable using only onboard sensors due to occlusion.
In a similar vein, \cite{liu2022regularized} considers the task of human-in-the-loop robot motion planning under full observability, with SDF representing the human body.
In contrast to considering a single entity such as a human, our framework considers a variety of complex objects encountered in an indoor scenario, and performs dynamic collision avoidance.
Furthermore, our framework operates under partial observability with a single RGB-D sensor, tackling a pragmatic limited field-of-view setting.

\section{Method}

\subsection{Problem Formulation}

We consider the setting where a mobile robot navigates towards a goal position $\mathbf{x}_g$ in its workspace $ \mathcal{X} \subset \mathbb{R}^3$.
The robot is equipped with a camera to sense a frustum of the environment via RGB-D frames $[ \mathbf{I}, \mathbf{D} ]$, while also provided access to its pose $\mathbf{p} \in \mathrm{\mathbf{SE}}(3)$ as proprioceptive information, at any time instance $t$.
Using this, we infer the signed distance field (SDF) to compute collision costs for a trajectory towards a goal position $\mathbf{x}_g \in \mathcal{X} \subset \mathbb{R}^3$, as well as updates to the trajectory which push it to collision-free workspace $\mathcal{X}_{free} = \mathcal{X} \setminus \mathcal{X}_{obs}$, where $\mathcal{X}_{obs}$ is the formidable workspace.
Let $\Omega: \mathcal{X} \mapsto \mathbb{R}$ denote the SDF of the environment.
Then, for any point $\mathbf{x} \in \mathcal{X}$, $\Omega(\mathbf{x}) \leq 0$ implies $\mathbf{x} \in \mathcal{X}_{obs}$ and $\nabla_{\mathbf{x}}\Omega(\mathbf{x})$ gives the direction in which the point can be pushed in order to exit $\mathcal{X}_{obs}$ with minimum travel.
Using this information, we propose a framework to infer SDF-based collision information from posed RGB-D frames to allow gradient-based trajectory optimization.

\subsection{Neural Signed Distance Functions}

Neural SDFs~\cite{deepsdf19,gropp2020implicit,isdf22} use neural networks to compactly represent the continuous SDF of the environment.
For dynamic environments, a straight forward approach to adapt neural SDFs is to repeatedly learn from data spanning the complete workspace.
This is computationally expensive, impractical (since maintaining data over the whole workspace is not scalable), and redundant (since most components in the scene do not change overtime).
Hence, we decompose the SDF inference procedure by separately modeling geometries at the object level and the scene level.
For this purpose, we use the following neural SDF formulations:

\begin{enumerate}
    \item \emph{Object-level SDFs}:
    To model SDFs $\{ \Omega_{o_i} \}_{i=1}^{N_o}$ of a catalog of objects $\mathcal{O} = \{o_i\}_{i=1}^{N_o}$, we use DeepSDF~\cite{deepsdf19}.
    It employs an auto-decoder formulation to learn the SDF of objects of varying shapes, in a compact neural network conditioned on a per-shape latent code.
    The signed distance is obtained as: $\Omega_{o_i} = \Omega_o (\cdot, z_i)$, where $z_i \in \mathbb{R}^d$ is the latent code for object $o_i$.
    \item \emph{Scene-level SDFs}:
    To model the SDF of the indoor scene $\Omega_s$ comprising \emph{static} background components (such as walls and floor), we use iSDF~\cite{isdf22} which can be trained from posed depth frames $\{(\mathbf{p}, \mathbf{D})\}$ in a continual learning framework.
\end{enumerate}

\noindent We train our neural SDFs offline, and their outputs are combined in online operation.
Therefore, the individual networks can be exchanged with alternatives in a plug-and-play fashion, resulting in a flexible, modular pipeline.

\subsection{\label{ssec:scene-sdf}Composite Scene SDF Pipeline}

\begin{figure}[t]
    \centering\vspace{3mm}
    \subfloat[\label{fig:scene-sdf}Scene SDF Pipeline]{
        \includegraphics[page=2, width=\columnwidth,trim={1.63cm, 16.4cm, 4.5cm, 7cm},clip]{figs/icra25-fig-pipeline.pdf}
    }
    \newline\vspace{-0.2in}
    \subfloat[\label{fig:robo-sdf}Robot SDF Pipeline]{
        \includegraphics[page=1, width=\columnwidth,trim={2.2cm, 17.2cm, 4.9cm, 7.8cm},clip]{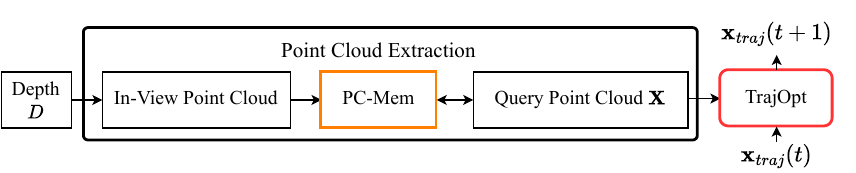}
    }
    \caption{\label{fig:scene-robo}
        Our proposed \emph{Dual Mode} pipeline comprises the two pipelines shown above.
        \textit{Scene SDF} pipeline (\ref{fig:scene-sdf}) infers the consolidated SDF of the robot's workspace by computing the mapping from the workspace to the domain of each individual SDF representation.
        On the other hand, the \textit{Robot SDF} pipeline (\ref{fig:robo-sdf}) directly operates on the visible scene's point cloud by querying the robot body's SDF along the robot trajectory.
    }
    \vspace{-3.5mm}  %
\end{figure}

Towards inferring the SDF of the visible scene, we present the Scene SDF pipeline shown in Fig.~\ref{fig:scene-sdf}.
We first infer the mapping between objects (in the workspace) and their SDF representations (in normalized coordinate space), and then compose them to generate the visible scene's SDF representation, which is queried to perform trajectory optimization.
The current visible point cloud $\mathbf{X}_v$ is extracted from the corresponding posed depth frame $(\mathbf{p}, \mathbf{D})$.
Next, points corresponding to the scene background are filtered-out using the trained scene SDF network $\Omega_s$.
The resultant point-set $\mathbf{X}_o = \{ \mathbf{x}^o \in \mathbf{X}_v \mid \Omega_s(\mathbf{x}^o) > 0 \}$ corresponds to the movable objects in the scene.
Using the RGB frame $\mathbf{I}$, an object detector generates bounding box detections $\{B_i^j\}$ (where $B_i^j$ corresponds to to the $j$-th instance of object $o_i$ in the scene: $o_i^j$), to which points in $\mathbf{X}_o$ are assigned in a hierarchical nearest-first fashion.
Resultantly, we obtain pairs of bounding boxes and point-sets $\{(B_i^j, \mathbf{X}_i^j)\}$.

\begin{figure*}[t]
    \centering\vspace{2mm}
    \includegraphics[page=1, width=\textwidth,trim={6.75cm, 3cm, 6.2cm, 2.5cm},clip]{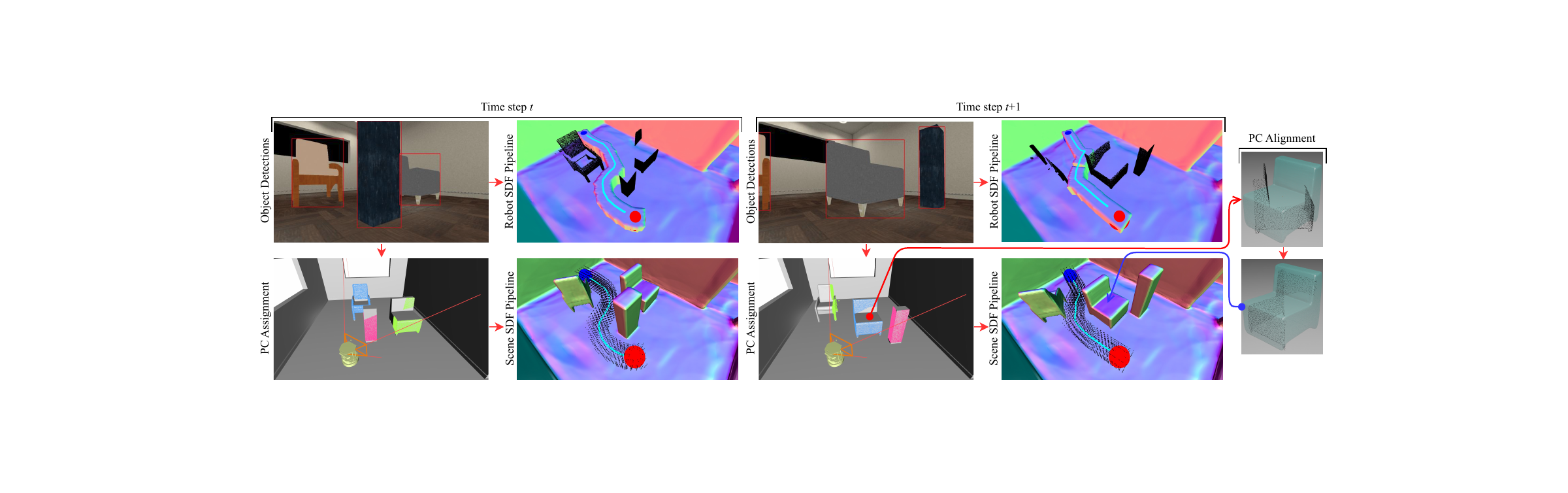}
    \caption{\label{fig:pipelines-dyn-steps}
        Execution of the proposed pipelines is demonstrated for a given object placement at time step $t$ and after a displacement at time step $t+1$ (shown in the respective \emph{PC Alignment} sections, where all three objects are displaced).
        The \emph{Robot SDF} pipeline queries the robot body's SDF at the obstacle point cloud (black) to guide trajectory optimization.
        The \emph{Scene SDF} pipeline assigns point cloud regions to each object instance, and then infers the mapping from the robot's workspace to the domain of each object's SDF representation.
        This mapping enables inferring the full SDF of the visible scene used to guide trajectory optimization.
        {\color{red} \textbf{Red}} and {\color{blue} \textbf{blue}} spheres denote robot's start and goal positions, respectively.
    }
    \vspace{-4mm}  %
\end{figure*}

For each detected object instance $o_i^j$ in the scene, we use $(B_i^j, \mathbf{X}_i^j)$ to compute a transformation $H_i^j$ which projects the points to the domain of the object's SDF $\Omega_{o_i}$.
Here, we optimize for plausible orientations only (e.g., a chair will never be upside down).
We start by computing a coarse estimate of the translation $T_i^j \in \mathbf{SE}(3)$ of the point-set $\mathbf{X}_i^j$, using its bounds in workspace, to bring it near origin.
Additionally, using prior information, we compute the scaling $S_i^j = \mathrm{diag}(\mathbf{s}) \in \mathbb{R}^{4\times4}$ from original object dimensions to the normalized object coordinate system, where $\mathbf{s} = [1, 1, 1, \gamma]$ and $\gamma > 0$ is the scaling factor.
Applying the transformations to the point-set $\mathbf{X}_i^j$ yields: $\bar{\mathbf{X}}_i^j = \{ S_i^j T_i^j \mathbf{x}^o \mid \mathbf{x}^o \in \mathbf{X}_i^j\}$.
We then align the point set $\bar{\mathbf{X}}_i^j$ with the object's SDF representation $\Omega_{o_i}$, such that $\Omega_{o_i}(R_i^j \bar{\mathbf{x}}^o ) \approx 0, \, \forall\, \bar{\mathbf{x}}^o \in \bar{\mathbf{X}}_i^j$ (shown in Fig.~\ref{fig:pipelines-dyn-steps}), by computing the transformation $R_i^j \in \mathbf{SE}(3)$ of the object $o_i^j$ using the following optimization objective function:

\begin{equation}
    \mathcal{L}_H(\bar{\mathbf{X}}_i^j) = \frac{1}{ | \bar{\mathbf{X}}_i^j | } \sum_{\bar{\mathbf{x}}^o \in \bar{\mathbf{X}}_i^j} \mathbb{I}_{\delta_{l}, \delta_{u}}(\Omega_{o_i}(R_i^j \bar{\mathbf{x}}^o))
\end{equation}

\noindent where $\mathbb{I}_{\delta_{l}, \delta_{u}}(v) = v$ if $v \in (\delta_{l}, \delta_{u})$, else $0$.
In practice, we start with $\delta_{l} = -0.2$ and $\delta_{u} = 0.01$ and progressively reduce the allowable region $(\delta_{l}, \delta_{u})$ in subsequent iterations to guide the optimization.
Consequently, we obtain a transformation $H_i^j = R_i^j S_i^j T_i^j$ such that $\forall\, \mathbf{x}^o \in \mathbf{X}_i^j$, $\Omega_{o_i}(H_i^j \mathbf{x}^o) \approx 0$.
Following our assumption of plausible object orientations, we optimize for translation and in-plane rotation only.

Performing the above procedure for each predicted object instance yields a set of tuples $\{ (\Omega_{o_i}, H_i^j) \}$, which we combine with the static scene SDF $\Omega_s$ to yield the full SDF $\Omega$ of the visible scene (as shown in Fig.~\ref{fig:pipelines-dyn-steps}):

\vspace{-3mm}
\begin{equation}
    \Omega(\mathbf{x}) = \mathrm{min}\Big( \Omega_s(\mathbf{x}), \mathrm{min}\big(\{\Omega_{o_i}(H_i^j \mathbf{x})\} \big) \Big), \,\,\, \mathbf{x} \in \mathbb{R}^3
\end{equation}

\noindent Note that $\Omega$ is fully differentiable, which we use to optimize an initial robot trajectory to goal $\mathbf{X}_r = \{ \mathbf{x}_i^r \}_{i=1}^{N_r}$ via the following cost function:

\vspace{-3mm}
\begin{align}
    \mathcal{L}_S(\mathbf{X}_r) = & \sum_{i=1}^{N_r}{ \frac{\alpha_i}{N_r} \mathrm{e}^{- \Omega(\mathbf{x}_i^r) } } + \lambda \sum_{i=1}^{N_r-1} \frac{(\mathbf{x}^r_{i+1} - \mathbf{x}^r_{i})^2}{N_r-1}
\end{align}

\noindent where $\lambda$ weighs-in path smoothing and $\alpha_i = \mathbb{I}(\Omega(x_i) < \zeta)$ enforces updates to occur only in the safety margin of $\zeta$.
In our experiments, we use $\lambda=0.1$ and $\zeta=0.05$, and use Adam optimizer.
In practice, $\Omega$ is evaluated on robot-sized oriented 3D grids centered at points along the trajectory.

\subsection{\label{ssec:robo-sdf}Robot Body SDF Formulation}

Instead of modeling the scene's SDF and querying it on the trajectory locations, we can opt for a simpler route:
model the robot's SDF and query it at obstacle point cloud locations~\cite{anyshapetrajopt23}, obtaining collision costs as well as gradients to perform trajectory optimization.
This procedure significantly reduces the modeling and inference costs while also working with fewer assumptions.
In this spirit, we implement the pipeline displayed in Fig.~\ref{fig:robo-sdf} which uses an SDF representation of the robot body's bounding box swept along the robot trajectory, as shown in Fig.~\ref{fig:pipelines-dyn-steps}.
We adopt trained object-level SDF network $\Omega_o$ and infer a latent code $\mathbf{z}_r$ to obtain the robot body's SDF $\Omega_r(\mathbf{x}) = \Omega_o(\mathbf{x}, \mathbf{z}_r)$.

To formulate the collision penalty for the complete trajectory, each point in the visible obstacle point cloud $\mathbf{X}_o$ is transformed to the domain of $\Omega_r$ centered at each point along the trajectory $\mathbf{X}_r$.
If $M_i^r \in \mathbb{R}^{4\times4}$ transforms points from the robot's workspace to the domain of $\Omega_r$ centered at trajectory point $\mathbf{x}_i^r$, then $\Omega_r(M_i^r \mathbf{x})$ gives the signed distance penalty imposed on trajectory point $\mathbf{x}_i^r$ by the point $\mathbf{x}$.

\vspace{-3mm}
\begin{align}
    \mathcal{L}_R(\mathbf{X}_r, \mathbf{X}_o) = & \sum_{i=1}^{N} \bigg( \sum_{\mathbf{x}^o \in \mathbf{X}_o}{ \frac{\alpha_i}{N*|\mathbf{X}_o|} \mathrm{e}^{- \Omega_r(M_i^r \mathbf{x}^o)} } \bigg) \notag\\
    & + \lambda \sum_{i=1}^{N - 1} \frac{ ( \mathbf{x}_{i+1}^r - \mathbf{x}_i^r )^2 }{N-1}
\end{align}

\noindent where $\alpha_i = \mathbb{I}(\Omega_r(M_i^r \mathbf{x}) < \zeta)$ enforce updates to occur only within the safety margin of $\zeta=0.05$, and $\lambda=0.1$ weighs-in path smoothing.
Furthermore, to reduce neural network inference cost, we only use a subset of the point cloud within a radius of $d_{traj}=0.5$m around the trajectory.
We use Adam optimizer to optimize our trajectories.

\subsection{Memory Modules}

Since the current objective constrains the task to a single ego-centric view setting, as the agent traverses the generated trajectory it may approach some obstacles too close to extract sufficient shape information, while looking away from other obstacles, forgetting to incorporate collision information in subsequent optimization steps.
Therefore, we resort to short-term memory modules for both stages of the pipeline.
For the Robot SDF pipeline, a \emph{Point Cloud Memory Module} (PC-Mem) persists out-of-view point clouds within a radius $r_{\mathrm{PC}}=1\mathrm{m}$ around the robot's position, which is updated with the robot's position and the camera's view frustum.

For the Scene SDF pipeline, an \emph{Obstacle Memory Module} (Obs-Mem) persists poses of nearby obstacles by dividing the workspace into three concentric sub-regions:
(a) Freeze zone $\mathcal{F}$, the immediate region around the robot ($\leq0.5\mathrm{m}$) where no new obstacle is considered,
(b) Update zone $\mathcal{U}$, the region immediately outside $\mathcal{F}$ ($\in \left(0.5\mathrm{m}, 1\mathrm{m}\right]$), where any encountered object will be memorized and if already in memory its cached pose information is used to reduce repeated computation, and
(c) \textit{Dynamic/Memoryless zone} $\mathcal{D}$ ($>1\mathrm{m}$), the region outside $\mathcal{U}$ where objects are not memorized.
Our memory modules provide a straightforward approach to tackle the limited field-of-view setting.

\begin{algorithm}[t]
    \caption{\label{algo:dual-mode}
        Dual Mode Pipeline for Trajectory Optimization
    }
    \begin{algorithmic}[1]
        \State{$F_t=[\mathbf{I}_t, \mathbf{D}_t, \mathbf{p}_t]$} \Comment{Input: posed RGB-D frame}
        \State{$\mathbf{x}_t$} \Comment{Input: previous trajectory}
        \State{$\mathbf{X}_v=\texttt{\textbf{Depth2PC}}(\mathbf{D}_t, \mathbf{p}_t)$} \Comment{Extract visible PC}
        \State{$\{B_i^j\}=\texttt{\textbf{ObjDet}}(\mathbf{I}_t)$} \Comment{Object detection}

        \State{$\mathbf{X}_o=\{ \mathbf{x} \in \mathbf{X}_v \mid \Omega_s(\mathbf{X}_v) > 0 \}$} \Comment{Extract object PCs}
        \State{$\mathbf{X} = \mathbf{X}_o \cup \textrm{PC-Mem}$}
        \State{$\mathbf{x}_{t'} = \texttt{\textbf{TrajOpt}}(\mathbf{x}_t, \mathbf{X}, \Omega_r)$} \Comment{Robot SDF}
        \State{$\textrm{PC-Mem}.\mathrm{update}(\mathbf{X}, \mathbf{p}_t)$} \Comment{Memory update}

        \If{$\texttt{\textbf{inCollision}}(\mathbf{x}_{t'}, \mathbf{X}, \Omega_r)$}
            \State{$ \{(B_i^j, \mathbf{X}_i^j)\} = \texttt{\textbf{PCAssign}}(\mathbf{X}_o, \{B_i^j\})$}  %
            \For{$o_i \in \mathcal{O}$}
                \State{$H_i^j = \texttt{\textbf{PCAlign}}(\Omega_{o_i}, B_i^j, \mathbf{X}_i^j) $} \Comment{Pose estimation}
                \State{$\mathrm{ObsMem}.\mathrm{update}(H_i^j, \mathbf{p}_t)$} \Comment{Memory update}
            \EndFor
            \State{$\mathcal{J} = \{ (\Omega_{o_i}, H_i^j) \} \cup \mathrm{ObsMem}$}
            \State{$\mathbf{x}_{t+1} = \texttt{\textbf{TrajOpt}}(\mathbf{x}_{t'}, \Omega_s, \mathcal{J})$} \Comment{Scene SDF}
        \Else
            \State{$\mathbf{x}_{t+1} = \mathbf{x}_{t'}$}
        \EndIf
        \State{\Return{$\mathbf{x}_{t+1}$}}  %
    \end{algorithmic}
\end{algorithm}

\subsection{Dual Mode Operation}

\begin{figure}[t]
    \centering\vspace{2mm}
    \includegraphics[width=\columnwidth, trim={1.25cm, 1.7cm, 1cm, 1.65cm},clip]{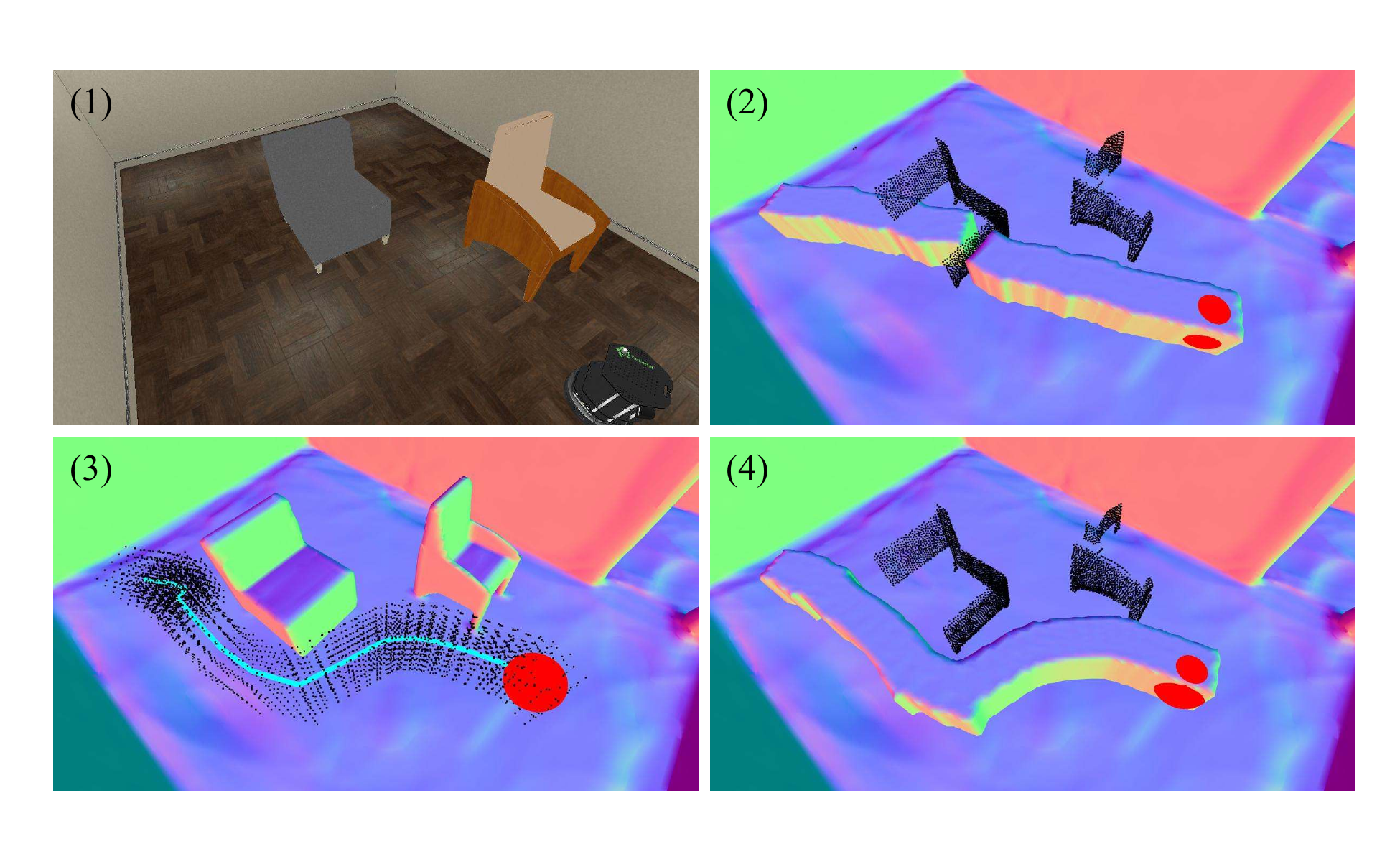}
    \caption{\label{fig:dual-mode-in-action}
        Operation of the \emph{Dual Mode} pipeline is shown in a simulated indoor scenario (1), where the trajectory from Robot Body SDF pipeline is stuck in a local minima (2), triggering the Scene SDF pipeline which generates a collision-free trajectory (3).
        Subsequently, the Robot Body SDF pipeline picks-up from this trajectory and continues operation (4).
    }
    \vspace{-3mm}
\end{figure}

The Robot Body SDF formulation of Sec.~\ref{ssec:robo-sdf} works well in most cases while requiring fewer assumptions as well as computation time.
When initialized in a collision-free region the Robot Body SDF pipeline sustains well for minor changes in collision costs as the scene evolves.
However, it is prone to getting trapped in local minima within the obstacle space, failing to generate a feasible trajectory, as shown in Fig.~\ref{fig:dual-mode-in-action}.
This can happen since we make use of a partial view of the scene at any point in time, yielding highly partial surface point clouds.
The Scene SDF pipeline of Sec.~\ref{ssec:scene-sdf} overcomes this issue by incorporating prior information, albeit incurring a higher computational cost and requiring a minimum distance assumption.
Furthermore, it may be redundant to frequently compute the visible scene's SDF.

Towards this, we propose a \emph{Dual Mode} pipeline, presented in Algorithm~\ref{algo:dual-mode}, which attempts to capitalize on the best of both worlds.
Our proposal uses the Robot Body SDF pipeline as the primary stage, performing the intuitive obstacle point cloud avoidance, and triggers the Scene SDF pipeline to infer the visible scene's SDF in case the primary stage fails.
This mode of operation enables more reliable trajectory optimization while significantly reducing the amortized computation time over the Scene SDF pipeline.
In effect, the trajectory generated by the Scene SDF pipeline acts as a better informed initialization for the Robot SDF's trajectory optimization procedure (shown in Fig.~\ref{fig:dual-mode-in-action}), which otherwise works well to quickly account for smaller changes in the scene.

\section{Experiments}

\subsection{Implementation Details}

We implement our pipeline in PyTorch and simulate an indoor scene with a Turtlebot4 in iGibson 2.0~\cite{igibson21}.
Obstacles comprise 80 different household objects.
During simulation, a random obstacle is selected for random displacement every 2 seconds.
The robot follows generated trajectories via a proportional controller for speed and a pure-pursuit controller~\cite{coulter1992purepursuit} for heading.
We use YOLOv5~\cite{yolov5} for object detection and implement our pipeline on an Intel Core i7 system with 32 GB RAM and an NVIDIA RTX 3090 GPU with 24GB VRAM.

\subsection{Baselines}

We compare our proposal with the following baselines:
\begin{enumerate}
    \item \textbf{\emph{EgoTrajOpt}}: EGO-Planner~\cite{zhou2020ego} improves efficiency over the standard SDF generation methods by computing a local approximation of the SDF which avoids taking into consideration the obstacles which do not contribute to trajectory optimization.
    We incorporate their distance field formulation in our PyTorch-based trajectory optimization framework as a benchmark.
    Similar to our proposal, this baseline uses ego-centric, single-view observations of the environment.
    \item \textbf{\emph{DWA+PC-Mem}}: We implement the Dynamic Window Approach~\cite{dwa97}, an online collision avoidance approach derived directly from the robot dynamics.
    Given the limited field of view, we couple DWA with the PC-Mem of Robot Body SDF pipeline.
    We set the planning horizon to 1 sec.
    \item \textbf{\emph{iRRT*}}: We implement Informed-RRT*~\cite{gammell2014informed}, a sampling-based planner which improves on the convergence rate of optimal RRT planners by focusing search in the sampling space to a prolate hyperspheroid.
    We provide the planner with full environment collision information.
    We set the upper limit for iterations to 300, with goal position sampled with probability $0.1$.
\end{enumerate}

\subsection{Quantitative Comparison}

\begin{table}[t]
    \centering\vspace{2mm}
    \hspace{-1mm}\scalebox{1}{
        \begin{tabular}{cccc}\toprule
            \multirow{2}{*}{Methods}&\multicolumn{3}{c}{Performance Metrics}\\ \cmidrule{2-4}
            &\multicolumn{1}{c}{S.R. $(\%)$ $\uparrow$}&\multicolumn{1}{c}{Traj. len (m) $\downarrow$}&\multicolumn{1}{c}{Plan time (s) $\downarrow$} \\\midrule

            \multirow{1}{*}{Dual Mode}& \multirow{1}{*}{$98.0$} & \multirow{1}{*}{$3.44 \pm 0.82$}  &\multirow{1}{*}{$15.73 \pm 6.99$} \\
            
            \multirow{1}{*}{Robot SDF}& \multirow{1}{*}{$96.33$} & \multirow{1}{*}{$3.43 \pm 0.82$}  &\multirow{1}{*}{$13.87 \pm 5.74$} \\
            
            \multirow{1}{*}{Scene SDF}& \multirow{1}{*}{$93.33$} & \multirow{1}{*}{$3.47 \pm 0.85$}  &\multirow{1}{*}{$55.97 \pm 22.75$} \\

            \multirow{1}{*}{EgoTrajOpt}& \multirow{1}{*}{$78.33$} & \multirow{1}{*}{$3.36 \pm 0.84$}  &\multirow{1}{*}{$72.32 \pm 43.99$} \\
            \multirow{1}{*}{DWA+PC-Mem}& \multirow{1}{*}{$85.67$} & \multirow{1}{*}{$3.68 \pm 1.81$}  &\multirow{1}{*}{$14.23 \pm 11.14$} \\
            \multirow{1}{*}{iRRT*}& \multirow{1}{*}{$80.33$} & \multirow{1}{*}{$3.40 \pm 0.83$}  &\multirow{1}{*}{$174.52 \pm 105.54$} \\
            \bottomrule
        \end{tabular}
    }
    \caption{\label{table:comparisons}
        Benchmark results in iGibson 2.0 are presented.
        Each method was tested on 300 simulations involving random start-goal positions and obstacles, and randomized obstacle displacements in-simulation.
        S.R. denotes success rate.
        Traj. len. is the average length of the path traversed by the robot.
        Plan time is the average of \emph{cumulative} (accrued) per-simulation plan times.
    }\vspace{-4mm}
\end{table}

\begin{figure*}[!h]
    \centering
    \subfloat{
        \hspace{-1mm}\includegraphics[width=0.79\columnwidth]{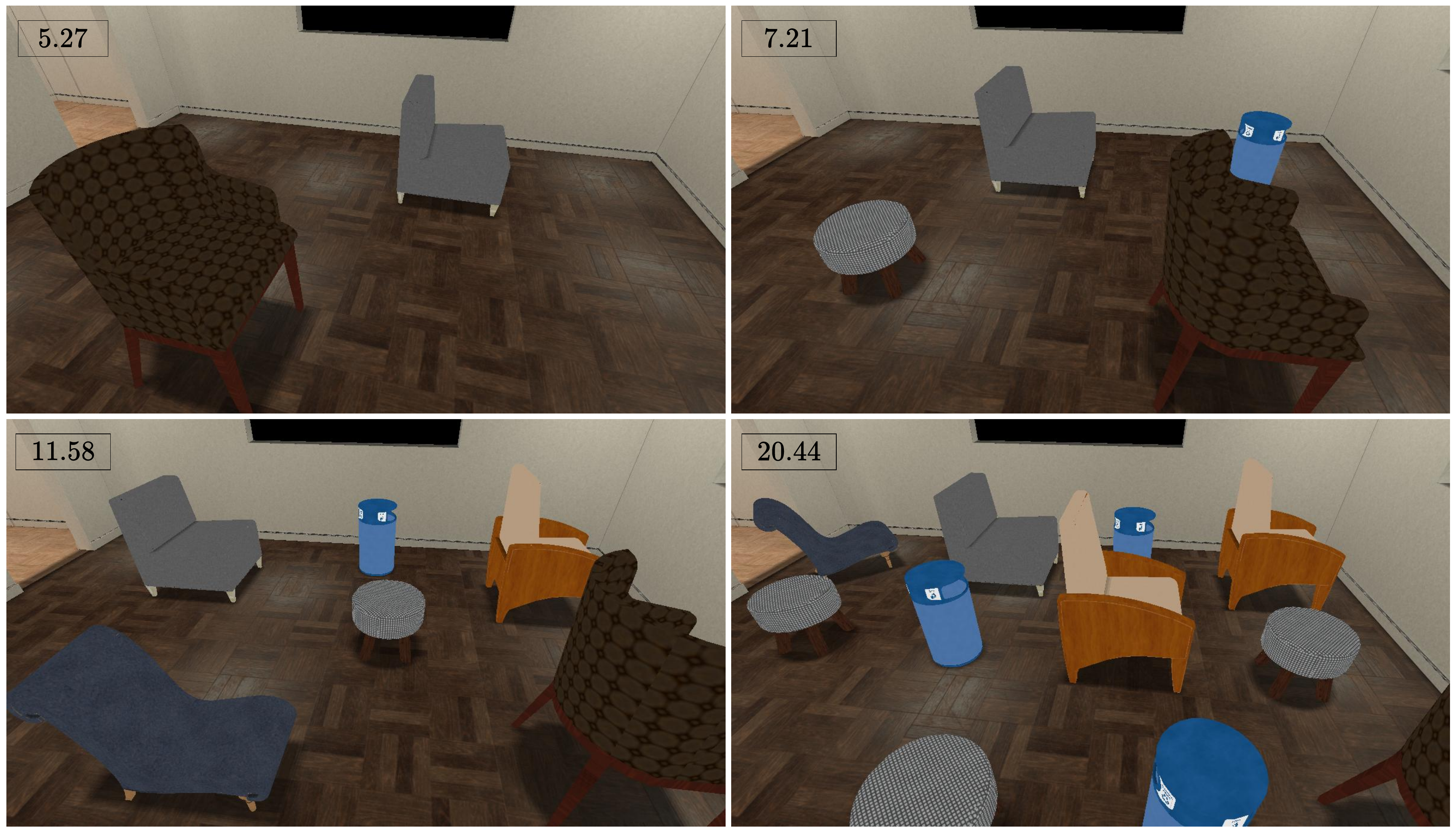}
    }
    \subfloat{
        \includegraphics[width=1.26\columnwidth,trim={1cm, 0.5cm, 1cm, 1.2cm},clip]{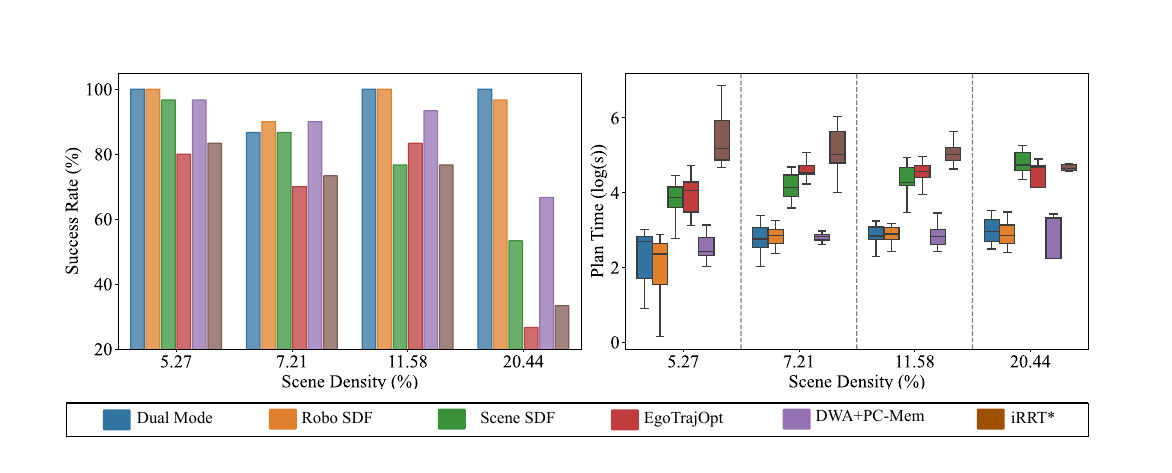}
    }
    \caption{\label{fig:sclb}
        We measure the impact of obstacle density in the robot's environment, using the scenes shown (\textit{left}, obstacle densities noted on top-left corners).
        Results (\textit{right}) show that the proposed \emph{Dual Mode} pipeline exhibits the lowest deterioration in performance as the scene density increases.
        Plan Times (corresponding to successful simulations only) are plotted on $\log$-scale for convenience.
    }
    \vspace{-4mm}
\end{figure*}

To evaluate the methods, we generated 10 different scenarios (initial obstacle placements) in iGibson 2.0, each with 10 different start and goal states for the robot, and performed each simulation thrice (to account for randomness).
Hence, each method is evaluated over 300 simulations, results for which are detailed in Table~\ref{table:comparisons}.
Our proposed \emph{Dual Mode} pipeline outperforms comparisons with a 95.67\% success rate while being competitive in plan times.
In comparison, the \emph{Robot SDF} and \emph{Scene SDF} pipelines under-perform individually, with the \emph{Scene SDF} also incurring a significantly larger plan time.
Hence, the amortization of planning cost reinforces our claim that re-computing the full scene's SDF at every time step is not necessary and we can usually sustain with the swift trajectory refinement offered by the \emph{Robot SDF} pipeline, while resorting to the \emph{Scene SDF} pipeline only when needed.

The \emph{EgoTrajOpt} baseline exhibits a lower success rate which we attribute to its distance field formulation, which is less stable for trajectory optimization without the additional safeguards described in \cite{zhou2020ego}, compared to the well-defined signed distance field formulation.
Furthermore, the \emph{EgoTrajOpt} baseline uses A* search to generate trajectories for in-collision segments, which does not serve as a good approximation for obstacle surface when the robot size is not insignificant.
Furthermore, it also ends up taking more steps to reach goal position, accruing plan time.
We note a similar issue in \emph{DWA+PC-Mem}, where the evaluation function is known to be sensitive to dynamic obstacles~\cite{dwa_impr_2024}, which at times causes the robot to go around a little more in order to reach destination, while also accruing plan times, as shown by a higher standard-deviation for trajectory lengths and cumulative plan times.
Lastly, \emph{iRRT*} not only incurs more planning time, it also fails more often due to the difficulty in sampling in a cluttered workspace.

\subsection{Scene Scalability Analysis}

To measure the effectiveness of each method in handling scene clutter, we evaluate the methods on scenes with higher variance in obstacle density (shown in Fig.~\ref{fig:sclb}), defined as: $|\mathcal{X}_{obs}| / |\mathcal{X}|$ (ratio of the size of formidable workspace region to that of the total workspace).
Our proposal of a \emph{Dual Mode} pipeline performs comparably or better than the next best baseline method, and within a reasonable time budget.
Among the rest, the \emph{Robot SDF} pipeline performs notably better since it only requires the visible point cloud and does not need to explicitly model the SDF of each object to compute collision costs.
This strategy persists its performance as scene clutter increases and full visibility of objects becomes more challenging.
\emph{Scene SDF} pipeline, being greatly dependent on visibility, suffers increasingly with scene density.

We note that as the scene clutter increases, so does the necessity for accurate collision costs and well-defined collision avoidance gradients, the lack of which attributes to the declining trend of the \emph{EgoTrajOpt} baseline.
\emph{DWA+PC-Mem} also demonstrates a declining trend for performance, however, we note that most failure cases correspond to breaching the simulation time limit for our experiments (20 minutes), which happens increasingly more as the scene clutter increases.
Furthermore, \emph{iRRT*} also shows a declining trend as increased clutter reduces collision-free sampling space, hence not yielding a collision-free plan within a reasonable time limit.
Hence, our proposal of a Dual Mode pipeline demonstrates its usefulness even as the environment clutter increases.

\subsection{Real-World Evaluation}

To deploy our pipeline in a real-world indoor environment, we interface with a Turtlebot4 platform, equipped with a RealSense 435i RGB-D camera, in ROS2~\cite{ros2}.
Depth data from the sensor is subject to noise, which exacerbates the difficulty of working with highly partial surface point clouds.
Hence, we work with the SDF of the convex hull of each object, and therefore, align point clouds to be encapsulated within the zero-level set.
This is a reasonable assumption given the overarching objective of motion planing with collision avoidance constraints.
Shown in Fig.~\ref{fig:teaser}, the robot navigates from one end of the room to another, while avoiding collisions with static (furniture adjacent to walls) as well as dynamic (the office chair and briefcase) scene components.
We provide more planning scenarios in the \emph{supplementary material}, including planning with human presence (which we do not model explicitly in our pipeline, however, the constituent \emph{Robot SDF} pipeline naturally handles).

\section{Conclusion}

We present a structured approach to vision-based robot navigation in dynamic indoor environments using neural signed distance fields.
Our proposal comprises two complementary components: A Scene SDF pipeline which models the visible scene's SDF by combining ego-centric visual input with prior knowledge, and a Robot Body SDF pipeline which models the robot's body to query on visible obstacle point cloud.
The components are complementary in nature, and enable more reliable trajectory generation with low amortized cost.
We evaluate our proposal in the iGibson simulated environment in the presence of clutter, and demonstrate its utility in real-world indoor scenarios.
The modularity of our structured approach allows the constituent components to easily be swapped with more efficient and performant alternatives in the future.

\bibliographystyle{IEEEtran}
\bibliography{IEEEabrv,references}

\begin{thebibliography}{10}
\providecommand{\url}[1]{#1}
\csname url@rmstyle\endcsname
\providecommand{\newblock}{\relax}
\providecommand{\bibinfo}[2]{#2}
\providecommand\BIBentrySTDinterwordspacing{\spaceskip=0pt\relax}
\providecommand\BIBentryALTinterwordstretchfactor{4}
\providecommand\BIBentryALTinterwordspacing{\spaceskip=\fontdimen2\font plus
\BIBentryALTinterwordstretchfactor\fontdimen3\font minus \fontdimen4\font\relax}
\providecommand\BIBforeignlanguage[2]{{%
\expandafter\ifx\csname l@#1\endcsname\relax
\typeout{** WARNING: IEEEtran.bst: No hyphenation pattern has been}%
\typeout{** loaded for the language `#1'. Using the pattern for}%
\typeout{** the default language instead.}%
\else
\language=\csname l@#1\endcsname
\fi
#2}}

\bibitem{finean2021predicted}
M.~N. Finean, W.~Merkt, and I.~Havoutis, ``Predicted composite signed-distance fields for real-time motion planning in dynamic environments,'' in \emph{Proceedings of the International Conference on Automated Planning and Scheduling}, vol.~31, 2021, pp. 616--624.

\bibitem{zucker2013chomp}
M.~Zucker, N.~Ratliff, A.~D. Dragan, M.~Pivtoraiko, M.~Klingensmith, C.~M. Dellin, J.~A. Bagnell, and S.~S. Srinivasa, ``Chomp: Covariant hamiltonian optimization for motion planning,'' \emph{The International journal of robotics research}, vol.~32, no. 9-10, pp. 1164--1193, 2013.

\bibitem{oleynikova2017voxblox}
H.~Oleynikova, Z.~Taylor, M.~Fehr, R.~Siegwart, and J.~Nieto, ``Voxblox: Incremental 3d euclidean signed distance fields for on-board mav planning,'' in \emph{2017 IEEE/RSJ International Conference on Intelligent Robots and Systems (IROS)}.\hskip 1em plus 0.5em minus 0.4em\relax IEEE, 2017, pp. 1366--1373.

\bibitem{deepsdf19}
J.~J. Park, P.~Florence, J.~Straub, R.~Newcombe, and S.~Lovegrove, ``Deepsdf: Learning continuous signed distance functions for shape representation,'' in \emph{The IEEE Conference on Computer Vision and Pattern Recognition (CVPR)}, June 2019.

\bibitem{isdf22}
J.~Ortiz, A.~Clegg, J.~Dong, E.~Sucar, D.~Novotny, M.~Zollhoefer, and M.~Mukadam, ``isdf: Real-time neural signed distance fields for robot perception,'' in \emph{Robotics: Science and Systems}, 2022.

\bibitem{gropp2020implicit}
A.~Gropp, L.~Yariv, N.~Haim, M.~Atzmon, and Y.~Lipman, ``Implicit geometric regularization for learning shapes,'' in \emph{International Conference on Machine Learning}.\hskip 1em plus 0.5em minus 0.4em\relax PMLR, 2020, pp. 3789--3799.

\bibitem{saulnier2020information}
K.~Saulnier, N.~Atanasov, G.~J. Pappas, and V.~Kumar, ``Information theoretic active exploration in signed distance fields,'' in \emph{2020 IEEE International Conference on Robotics and Automation (ICRA)}.\hskip 1em plus 0.5em minus 0.4em\relax IEEE, 2020, pp. 4080--4085.

\bibitem{hiosdf2023}
V.~Vasilopoulos, S.~Garg, J.~Huh, B.~Lee, and V.~Isler, ``Hio-sdf: Hierarchical incremental online signed distance fields,'' in \emph{2024 IEEE International Conference on Robotics and Automation (ICRA)}.\hskip 1em plus 0.5em minus 0.4em\relax IEEE, 2024, pp. 17\,537--17\,543.

\bibitem{nfomp2022}
M.~Kurenkov, A.~Potapov, A.~Savinykh, E.~Yudin, E.~Kruzhkov, P.~Karpyshev, and D.~Tsetserukou, ``Nfomp: Neural field for optimal motion planner of differential drive robots with nonholonomic constraints,'' \emph{IEEE Robotics and Automation Letters}, vol.~7, no.~4, pp. 10\,991--10\,998, 2022.

\bibitem{desouza2002vision}
G.~N. DeSouza and A.~C. Kak, ``Vision for mobile robot navigation: A survey,'' \emph{IEEE transactions on pattern analysis and machine intelligence}, vol.~24, no.~2, pp. 237--267, 2002.

\bibitem{cox1991blanche}
I.~J. Cox, ``Blanche-an experiment in guidance and navigation of an autonomous robot vehicle,'' \emph{IEEE Transactions on robotics and automation}, vol.~7, no.~2, pp. 193--204, 1991.

\bibitem{burgard1999experiences}
W.~Burgard, A.~B. Cremers, D.~Fox, D.~H{\"a}hnel, G.~Lakemeyer, D.~Schulz, W.~Steiner, and S.~Thrun, ``Experiences with an interactive museum tour-guide robot,'' \emph{Artificial intelligence}, vol. 114, no. 1-2, pp. 3--55, 1999.

\bibitem{thrun1999minerva}
S.~Thrun, M.~Bennewitz, W.~Burgard, A.~B. Cremers, F.~Dellaert, D.~Fox, D.~Hahnel, C.~Rosenberg, N.~Roy, J.~Schulte, \emph{et~al.}, ``Minerva: A second-generation museum tour-guide robot,'' in \emph{Proceedings 1999 IEEE International Conference on Robotics and Automation (Cat. No. 99CH36288C)}, vol.~3.\hskip 1em plus 0.5em minus 0.4em\relax IEEE, 1999.

\bibitem{karaman2011sampling}
S.~Karaman and E.~Frazzoli, ``Sampling-based algorithms for optimal motion planning,'' \emph{The international journal of robotics research}, vol.~30, no.~7, pp. 846--894, 2011.

\bibitem{jones2011visual}
E.~S. Jones and S.~Soatto, ``Visual-inertial navigation, mapping and localization: A scalable real-time causal approach,'' \emph{The International Journal of Robotics Research}, vol.~30, no.~4, pp. 407--430, 2011.

\bibitem{chaplot2020learning}
D.~S. Chaplot, D.~Gandhi, S.~Gupta, A.~Gupta, and R.~Salakhutdinov, ``Learning to explore using active neural slam,'' in \emph{International Conference on Learning Representations}, 2020.

\bibitem{zhan2022activermap}
H.~Zhan, J.~Zheng, Y.~Xu, I.~Reid, and H.~Rezatofighi, ``Activermap: Radiance field for active mapping and planning,'' \emph{arXiv preprint arXiv:2211.12656}, 2022.

\bibitem{gupta2017cognitive}
S.~Gupta, J.~Davidson, S.~Levine, R.~Sukthankar, and J.~Malik, ``Cognitive mapping and planning for visual navigation,'' in \emph{Proceedings of the IEEE conference on computer vision and pattern recognition}, 2017, pp. 2616--2625.

\bibitem{finn2017one}
C.~Finn, T.~Yu, T.~Zhang, P.~Abbeel, and S.~Levine, ``One-shot visual imitation learning via meta-learning,'' in \emph{Conference on robot learning}.\hskip 1em plus 0.5em minus 0.4em\relax PMLR, 2017, pp. 357--368.

\bibitem{stepputtis2020language}
S.~Stepputtis, J.~Campbell, M.~Phielipp, S.~Lee, C.~Baral, and H.~Ben~Amor, ``Language-conditioned imitation learning for robot manipulation tasks,'' \emph{Advances in Neural Information Processing Systems}, vol.~33, pp. 13\,139--13\,150, 2020.

\bibitem{zhu2017target}
Y.~Zhu, R.~Mottaghi, E.~Kolve, J.~J. Lim, A.~Gupta, L.~Fei-Fei, and A.~Farhadi, ``Target-driven visual navigation in indoor scenes using deep reinforcement learning,'' in \emph{2017 IEEE international conference on robotics and automation (ICRA)}.\hskip 1em plus 0.5em minus 0.4em\relax IEEE, 2017, pp. 3357--3364.

\bibitem{mirowski2016learning}
P.~Mirowski, R.~Pascanu, F.~Viola, H.~Soyer, A.~Ballard, A.~Banino, M.~Denil, R.~Goroshin, L.~Sifre, K.~Kavukcuoglu, \emph{et~al.}, ``Learning to navigate in complex environments,'' in \emph{International Conference on Learning Representations}, 2016.

\bibitem{ye2021auxiliary}
J.~Ye, D.~Batra, A.~Das, and E.~Wijmans, ``Auxiliary tasks and exploration enable objectgoal navigation,'' in \emph{Proceedings of the IEEE/CVF International Conference on Computer Vision}, 2021, pp. 16\,117--16\,126.

\bibitem{oleynikova2016signed}
H.~Oleynikova, A.~Millane, Z.~Taylor, E.~Galceran, J.~Nieto, and R.~Siegwart, ``Signed distance fields: A natural representation for both mapping and planning,'' in \emph{RSS 2016 workshop: geometry and beyond-representations, physics, and scene understanding for robotics}.\hskip 1em plus 0.5em minus 0.4em\relax University of Michigan, 2016.

\bibitem{mildenhall2021nerf}
B.~Mildenhall, P.~P. Srinivasan, M.~Tancik, J.~T. Barron, R.~Ramamoorthi, and R.~Ng, ``Nerf: Representing scenes as neural radiance fields for view synthesis,'' \emph{Communications of the ACM}, vol.~65, no.~1, pp. 99--106, 2021.

\bibitem{ni2022ntfields}
R.~Ni and A.~H. Qureshi, ``Ntfields: Neural time fields for physics-informed robot motion planning,'' in \emph{The Eleventh International Conference on Learning Representations}, 2022.

\bibitem{pan2022voxfield}
Y.~Pan, Y.~Kompis, L.~Bartolomei, R.~Mascaro, C.~Stachniss, and M.~Chli, ``Voxfield: Non-projective signed distance fields for online planning and 3d reconstruction,'' in \emph{2022 IEEE/RSJ International Conference on Intelligent Robots and Systems (IROS)}.\hskip 1em plus 0.5em minus 0.4em\relax IEEE, 2022, pp. 5331--5338.

\bibitem{han2019fiesta}
L.~Han, F.~Gao, B.~Zhou, and S.~Shen, ``Fiesta: Fast incremental euclidean distance fields for online motion planning of aerial robots,'' in \emph{2019 IEEE/RSJ International Conference on Intelligent Robots and Systems (IROS)}.\hskip 1em plus 0.5em minus 0.4em\relax IEEE, 2019, pp. 4423--4430.

\bibitem{zhou2020ego}
X.~Zhou, Z.~Wang, H.~Ye, C.~Xu, and F.~Gao, ``Ego-planner: An esdf-free gradient-based local planner for quadrotors,'' \emph{IEEE Robotics and Automation Letters}, vol.~6, no.~2, pp. 478--485, 2020.

\bibitem{yang2023iplanner}
F.~Yang, C.~Wang, C.~Cadena, and M.~Hutter, ``iplanner: Imperative path planning,'' in \emph{Robotics: Science and Systems}, 2023.

\bibitem{adamkiewicz2022vision}
M.~Adamkiewicz, T.~Chen, A.~Caccavale, R.~Gardner, P.~Culbertson, J.~Bohg, and M.~Schwager, ``Vision-only robot navigation in a neural radiance world,'' \emph{IEEE Robotics and Automation Letters}, vol.~7, no.~2, pp. 4606--4613, 2022.

\bibitem{liu2022regularized}
P.~Liu, K.~Zhang, D.~Tateo, S.~Jauhri, J.~Peters, and G.~Chalvatzaki, ``Regularized deep signed distance fields for reactive motion generation,'' in \emph{2022 IEEE/RSJ International Conference on Intelligent Robots and Systems (IROS)}.\hskip 1em plus 0.5em minus 0.4em\relax IEEE, 2022, pp. 6673--6680.

\bibitem{anyshapetrajopt23}
T.~Zhang, J.~Wang, C.~Xu, A.~Gao, and F.~Gao, ``Continuous implicit sdf based any-shape robot trajectory optimization,'' in \emph{2023 IEEE/RSJ International Conference on Intelligent Robots and Systems (IROS)}.\hskip 1em plus 0.5em minus 0.4em\relax IEEE, 2023, pp. 282--289.

\bibitem{igibson21}
C.~Li, F.~Xia, R.~Mart{\'\i}n-Mart{\'\i}n, M.~Lingelbach, S.~Srivastava, B.~Shen, K.~E. Vainio, C.~Gokmen, G.~Dharan, T.~Jain, \emph{et~al.}, ``igibson 2.0: Object-centric simulation for robot learning of everyday household tasks,'' in \emph{Conference on Robot Learning}.\hskip 1em plus 0.5em minus 0.4em\relax PMLR, 2022, pp. 455--465.

\bibitem{coulter1992purepursuit}
R.~C. Coulter \emph{et~al.}, \emph{Implementation of the pure pursuit path tracking algorithm}.\hskip 1em plus 0.5em minus 0.4em\relax Carnegie Mellon University, The Robotics Institute, 1992.

\bibitem{yolov5}
\BIBentryALTinterwordspacing
G.~Jocher, ``Yolov5 by ultralytics,'' 2020. [Online]. Available: \url{https://github.com/ultralytics/yolov5}
\BIBentrySTDinterwordspacing

\bibitem{dwa97}
D.~Fox, W.~Burgard, and S.~Thrun, ``The dynamic window approach to collision avoidance,'' \emph{IEEE Robotics \& Automation Magazine}, vol.~4, no.~1, pp. 23--33, 1997.

\bibitem{gammell2014informed}
J.~D. Gammell, S.~S. Srinivasa, and T.~D. Barfoot, ``Informed rrt: Optimal sampling-based path planning focused via direct sampling of an admissible ellipsoidal heuristic,'' in \emph{2014 IEEE/RSJ international conference on intelligent robots and systems}.\hskip 1em plus 0.5em minus 0.4em\relax IEEE, 2014, pp. 2997--3004.

\bibitem{dwa_impr_2024}
Y.~Cao and N.~M. Nor, ``An improved dynamic window approach algorithm for dynamic obstacle avoidance in mobile robot formation,'' \emph{Decision Analytics Journal}, vol.~11, p. 100471, 2024.

\bibitem{ros2}
\BIBentryALTinterwordspacing
S.~Macenski, T.~Foote, B.~Gerkey, C.~Lalancette, and W.~Woodall, ``Robot operating system 2: Design, architecture, and uses in the wild,'' \emph{Science Robotics}, vol.~7, no.~66, p. eabm6074, 2022. [Online]. Available: \url{https://www.science.org/doi/abs/10.1126/scirobotics.abm6074}
\BIBentrySTDinterwordspacing

\end{thebibliography}

\end{document}